\batchmode
\makeatletter
\def\input@path{{/home/fyzhu/DATA/Dropbox/self_Folder/myWorksOnDropboxs/201702_IJCAI_RobustContextualBanit/}}
\makeatother
\documentclass[letterpaper]{article}
\usepackage{helvet}
\usepackage{courier}
\usepackage[latin9]{inputenc}
\usepackage{color}
\definecolor{page_backgroundcolor}{rgb}{1, 1, 1}
\pagecolor{page_backgroundcolor}
\usepackage{float}
\usepackage{bm}
\usepackage{amsmath}
\usepackage{amsthm}
\usepackage{graphicx}
\usepackage[unicode=true,pdfusetitle,
 bookmarks=true,bookmarksnumbered=false,bookmarksopen=false,
 breaklinks=true,pdfborder={0 0 0},pdfborderstyle={},backref=false,colorlinks=true]
 {hyperref}
\hypersetup{
 citecolor=blue, urlcolor=blue, linkcolor=blue}

\makeatletter

\pdfpageheight\paperheight
\pdfpagewidth\paperwidth

\floatstyle{ruled}
\newfloat{algorithm}{tbp}{loa}
\providecommand{\algorithmname}{Algorithm}
\floatname{algorithm}{\protect\algorithmname}

\theoremstyle{plain}
\newtheorem{thm}{\protect\theoremname}
\theoremstyle{plain}
\newtheorem{prop}[thm]{\protect\propositionname}
\theoremstyle{plain}
\newtheorem{lem}[thm]{\protect\lemmaname}
\ifx\proof\undefined
\newenvironment{proof}[1][\protect\proofname]{\par
\normalfont\topsep6\p@\@plus6\p@\relax
\trivlist
\itemindent\parindent
\item[\hskip\labelsep\scshape #1]\ignorespaces
}{%
\endtrivlist\@endpefalse
}
\providecommand{\proofname}{Proof}
\fi
\theoremstyle{remark}
\newtheorem{rem}[thm]{\protect\remarkname}

\usepackage{aaai18}
\usepackage{times}
\usepackage{url}
\usepackage{breakurl}
\usepackage{epsfig}
\usepackage{multirow}
\usepackage{mdwlist}
\usepackage{algorithm}
\usepackage{algorithmic}%
\usepackage{enumitem}
\usepackage{array}
\usepackage{float}
\usepackage{amsmath,amsthm,amssymb}
\usepackage{amsfonts}

\makeatother

\providecommand{\lemmaname}{Lemma}
\providecommand{\propositionname}{Proposition}
\providecommand{\remarkname}{Remark}
\providecommand{\theoremname}{Theorem}

\begin{document}
\global\long\def\mtbfA{\mathbf{A}}
 \global\long\def\mtbfa{\mathbf{a}}
 \global\long\def\mebfA{\bar{\mtbfA}}
 \global\long\def\mebfa{\bar{\mtbfa}}

\global\long\def\mhbfA{\widehat{\mathbf{A}}}
 \global\long\def\mhbfa{\widehat{\mathbf{a}}}
 \global\long\def\mtcalA{\mathcal{A}}
 \global\long\def\mtbbA{\mathbb{A}}

\global\long\def\mtbfB{\mathbf{B}}
 \global\long\def\mtbfb{\mathbf{b}}
 \global\long\def\mebfB{\bar{\mtbfB}}
 \global\long\def\mebfb{\bar{\mtbfb}}

\global\long\def\mhbfB{\widehat{\mathbf{B}}}
 \global\long\def\mhbfb{\widehat{\mathbf{b}}}
 \global\long\def\mtcalB{\mathcal{B}}
 \global\long\def\mtbbB{\mathbb{B}}

\global\long\def\mtbfC{\mathbf{C}}
 \global\long\def\mtbfc{\mathbf{c}}
 \global\long\def\mebfC{\bar{\mtbfC}}
 \global\long\def\mebfc{\bar{\mtbfc}}

\global\long\def\mhbfC{\widehat{\mathbf{C}}}
 \global\long\def\mhbfc{\widehat{\mathbf{c}}}
 \global\long\def\mtcalC{\mathcal{C}}
 \global\long\def\mtbbC{\mathbb{C}}

\global\long\def\mtbfD{\mathbf{D}}
 \global\long\def\mtbfd{\mathbf{d}}
 \global\long\def\mebfD{\bar{\mtbfD}}
 \global\long\def\mebfd{\bar{\mtbfd}}

\global\long\def\mhbfD{\widehat{\mathbf{D}}}
 \global\long\def\mhbfd{\widehat{\mathbf{d}}}
 \global\long\def\mtcalD{\mathcal{D}}
 \global\long\def\mtbbD{\mathbb{D}}

\global\long\def\mtbfE{\mathbf{E}}
 \global\long\def\mtbfe{\mathbf{e}}
 \global\long\def\mebfE{\bar{\mtbfE}}
 \global\long\def\mebfe{\bar{\mtbfe}}

\global\long\def\mhbfE{\widehat{\mathbf{E}}}
 \global\long\def\mhbfe{\widehat{\mathbf{e}}}
 \global\long\def\mtcalE{\mathcal{E}}
 \global\long\def\mtbbE{\mathbb{E}}

\global\long\def\mtbfF{\mathbf{F}}
 \global\long\def\mtbff{\mathbf{f}}
 \global\long\def\mebfF{\bar{\mathbf{F}}}
 \global\long\def\mebff{\bar{\mathbf{f}}}

\global\long\def\mhbfF{\widehat{\mathbf{F}}}
 \global\long\def\mhbff{\widehat{\mathbf{f}}}
 \global\long\def\mtcalF{\mathcal{F}}
 \global\long\def\mtbbF{\mathbb{F}}

\global\long\def\mtbfG{\mathbf{G}}
 \global\long\def\mtbfg{\mathbf{g}}
 \global\long\def\mebfG{\bar{\mathbf{G}}}
 \global\long\def\mebfg{\bar{\mathbf{g}}}

\global\long\def\mhbfG{\widehat{\mathbf{G}}}
 \global\long\def\mhbfg{\widehat{\mathbf{g}}}
 \global\long\def\mtcalG{\mathcal{G}}
 \global\long\def\mtbbG{\mathbb{G}}

\global\long\def\mtbfH{\mathbf{H}}
 \global\long\def\mtbfh{\mathbf{h}}
 \global\long\def\mebfH{\bar{\mathbf{H}}}
 \global\long\def\mebfh{\bar{\mathbf{h}}}

\global\long\def\mhbfH{\widehat{\mathbf{H}}}
 \global\long\def\mhbfh{\widehat{\mathbf{h}}}
 \global\long\def\mtcalH{\mathcal{H}}
 \global\long\def\mtbbH{\mathbb{H}}

\global\long\def\mtbfI{\mathbf{I}}
 \global\long\def\mtbfi{\mathbf{i}}
 \global\long\def\mebfI{\bar{\mathbf{I}}}
 \global\long\def\mebfi{\bar{\mathbf{i}}}

\global\long\def\mhbfI{\widehat{\mathbf{I}}}
 \global\long\def\mhbfi{\widehat{\mathbf{i}}}
 \global\long\def\mtcalI{\mathcal{I}}
 \global\long\def\mtbbI{\mathbb{I}}

\global\long\def\mtbfJ{\mathbf{J}}
 \global\long\def\mtbfj{\mathbf{j}}
 \global\long\def\mebfJ{\bar{\mathbf{J}}}
 \global\long\def\mebfj{\bar{\mathbf{j}}}

\global\long\def\mhbfJ{\widehat{\mathbf{J}}}
 \global\long\def\mhbfj{\widehat{\mathbf{j}}}
 \global\long\def\mtcalJ{\mathcal{J}}
 \global\long\def\mtbbJ{\mathbb{J}}

\global\long\def\mtbfK{\mathbf{K}}
 \global\long\def\mtbfk{\mathbf{k}}
 \global\long\def\mebfK{\bar{\mathbf{K}}}
 \global\long\def\mebfk{\bar{\mathbf{k}}}

\global\long\def\mhbfK{\widehat{\mathbf{K}}}
 \global\long\def\mhbfk{\widehat{\mathbf{k}}}
 \global\long\def\mtcalK{\mathcal{K}}
 \global\long\def\mtbbK{\mathbb{K}}

\global\long\def\mtbfL{\mathbf{L}}
 \global\long\def\mtbfl{\mathbf{l}}
 \global\long\def\mebfL{\bar{\mathbf{L}}}
 \global\long\def\mebfl{\bar{\mathbf{l}}}

\global\long\def\mhbfL{\widehat{\mathbf{K}}}
 \global\long\def\mhbfl{\widehat{\mathbf{k}}}
 \global\long\def\mtcalL{\mathcal{L}}
 \global\long\def\mtbbL{\mathbb{L}}

\global\long\def\mtbfM{\mathbf{M}}
 \global\long\def\mtbfm{\mathbf{m}}
 \global\long\def\mebfM{\bar{\mathbf{M}}}
 \global\long\def\mebfm{\bar{\mathbf{m}}}

\global\long\def\mhbfM{\widehat{\mathbf{M}}}
 \global\long\def\mhbfm{\widehat{\mathbf{m}}}
 \global\long\def\mtcalM{\mathcal{M}}
 \global\long\def\mtbbM{\mathbb{M}}

\global\long\def\mtbfN{\mathbf{N}}
 \global\long\def\mtbfn{\mathbf{n}}
 \global\long\def\mebfN{\bar{\mathbf{N}}}
 \global\long\def\mebfn{\bar{\mathbf{n}}}

\global\long\def\mhbfN{\widehat{\mathbf{N}}}
 \global\long\def\mhbfn{\widehat{\mathbf{n}}}
 \global\long\def\mtcalN{\mathcal{N}}
 \global\long\def\mtbbN{\mathbb{N}}

\global\long\def\mtbfO{\mathbf{O}}
 \global\long\def\mtbfo{\mathbf{o}}
 \global\long\def\mebfO{\bar{\mathbf{O}}}
 \global\long\def\mebfo{\bar{\mathbf{o}}}

\global\long\def\mhbfO{\widehat{\mathbf{O}}}
 \global\long\def\mhbfo{\widehat{\mathbf{o}}}
 \global\long\def\mtcalO{\mathcal{O}}
 \global\long\def\mtbbO{\mathbb{O}}

\global\long\def\mtbfP{\mathbf{P}}
 \global\long\def\mtbfp{\mathbf{p}}
 \global\long\def\mebfP{\bar{\mathbf{P}}}
 \global\long\def\mebfp{\bar{\mathbf{p}}}

\global\long\def\mhbfP{\widehat{\mathbf{P}}}
 \global\long\def\mhbfp{\widehat{\mathbf{p}}}
 \global\long\def\mtcalP{\mathcal{P}}
 \global\long\def\mtbbP{\mathbb{P}}

\global\long\def\mtbfQ{\mathbf{Q}}
 \global\long\def\mtbfq{\mathbf{q}}
 \global\long\def\mebfQ{\bar{\mathbf{Q}}}
 \global\long\def\mebfq{\bar{\mathbf{q}}}

\global\long\def\mhbfQ{\widehat{\mathbf{Q}}}
 \global\long\def\mhbfq{\widehat{\mathbf{q}}}
\global\long\def\mtcalQ{\mathcal{Q}}
 \global\long\def\mtbbQ{\mathbb{Q}}

\global\long\def\mtbfR{\mathbf{R}}
 \global\long\def\mtbfr{\mathbf{r}}
 \global\long\def\mebfR{\bar{\mathbf{R}}}
 \global\long\def\mebfr{\bar{\mathbf{r}}}

\global\long\def\mhbfR{\widehat{\mathbf{R}}}
 \global\long\def\mhbfr{\widehat{\mathbf{r}}}
\global\long\def\mtcalR{\mathcal{R}}
 \global\long\def\mtbbR{\mathbb{R}}

\global\long\def\mtbfS{\mathbf{S}}
 \global\long\def\mtbfs{\mathbf{s}}
 \global\long\def\mebfS{\bar{\mathbf{S}}}
 \global\long\def\mebfs{\bar{\mathbf{s}}}

\global\long\def\mhbfS{\widehat{\mathbf{S}}}
 \global\long\def\mhbfs{\widehat{\mathbf{s}}}
\global\long\def\mtcalS{\mathcal{S}}
 \global\long\def\mtbbS{\mathbb{S}}

\global\long\def\mtbfT{\mathbf{T}}
 \global\long\def\mtbft{\mathbf{t}}
 \global\long\def\mebfT{\bar{\mathbf{T}}}
 \global\long\def\mebft{\bar{\mathbf{t}}}

\global\long\def\mhbfT{\widehat{\mathbf{T}}}
 \global\long\def\mhbft{\widehat{\mathbf{t}}}
 \global\long\def\mtcalT{\mathcal{T}}
 \global\long\def\mtbbT{\mathbb{T}}

\global\long\def\mtbfU{\mathbf{U}}
 \global\long\def\mtbfu{\mathbf{u}}
 \global\long\def\mebfU{\bar{\mathbf{U}}}
 \global\long\def\mebfu{\bar{\mathbf{u}}}

\global\long\def\mhbfU{\widehat{\mathbf{U}}}
 \global\long\def\mhbfu{\widehat{\mathbf{u}}}
 \global\long\def\mtcalU{\mathcal{U}}
 \global\long\def\mtbbU{\mathbb{U}}

\global\long\def\mtbfV{\mathbf{V}}
 \global\long\def\mtbfv{\mathbf{v}}
 \global\long\def\mebfV{\bar{\mathbf{V}}}
 \global\long\def\mebfv{\bar{\mathbf{v}}}

\global\long\def\mhbfV{\widehat{\mathbf{V}}}
 \global\long\def\mhbfv{\widehat{\mathbf{v}}}
\global\long\def\mtcalV{\mathcal{V}}
 \global\long\def\mtbbV{\mathbb{V}}

\global\long\def\mtbfW{\mathbf{W}}
 \global\long\def\mtbfw{\mathbf{w}}
 \global\long\def\mebfW{\bar{\mathbf{W}}}
 \global\long\def\mebfw{\bar{\mathbf{w}}}

\global\long\def\mhbfW{\widehat{\mathbf{W}}}
 \global\long\def\mhbfw{\widehat{\mathbf{w}}}
 \global\long\def\mtcalW{\mathcal{W}}
 \global\long\def\mtbbW{\mathbb{W}}

\global\long\def\mtbfX{\mathbf{X}}
 \global\long\def\mtbfx{\mathbf{x}}
 \global\long\def\mebfX{\bar{\mtbfX}}
 \global\long\def\mebfx{\bar{\mtbfx}}

\global\long\def\mhbfX{\widehat{\mathbf{X}}}
 \global\long\def\mhbfx{\widehat{\mathbf{x}}}
 \global\long\def\mtcalX{\mathcal{X}}
 \global\long\def\mtbbX{\mathbb{X}}

\global\long\def\mtbfY{\mathbf{Y}}
 \global\long\def\mtbfy{\mathbf{y}}
\global\long\def\mebfY{\bar{\mathbf{Y}}}
 \global\long\def\mebfy{\bar{\mathbf{y}}}

\global\long\def\mhbfY{\widehat{\mathbf{Y}}}
 \global\long\def\mhbfy{\widehat{\mathbf{y}}}
 \global\long\def\mtcalY{\mathcal{Y}}
 \global\long\def\mtbbY{\mathbb{Y}}

\global\long\def\mtbfZ{\mathbf{Z}}
 \global\long\def\mtbfz{\mathbf{z}}
 \global\long\def\mebfZ{\bar{\mathbf{Z}}}
 \global\long\def\mebfz{\bar{\mathbf{z}}}

\global\long\def\mhbfZ{\widehat{\mathbf{Z}}}
 \global\long\def\mhbfz{\widehat{\mathbf{z}}}
\global\long\def\mtcalZ{\mathcal{Z}}
 \global\long\def\mtbbZ{\mathbb{Z}}

\global\long\def\mtth{\text{th}}

\global\long\def\mtbfzero{\mathbf{0}}
 \global\long\def\mtbfone{\mathbf{1}}

\global\long\def\mttrace{\text{Tr}}

\global\long\def\mttotalVariation{\text{TV}}

\global\long\def\mtexpect{\mathbb{E}}

\global\long\def\mtdet{\text{det}}

\global\long\def\mtvec{\mathbf{\text{vec}}}

\global\long\def\mtvar{\mathbf{\text{var}}}

\global\long\def\mtcov{\mathbf{\text{cov}}}

\global\long\def\mtsubTo{\mathbf{\text{s.t.}}}

\global\long\def\mtfor{\text{for}}

\global\long\def\mtrank{\text{rank}}

\global\long\def\mtrankn{\text{rankn}}

\global\long\def\mtdiag{\mathbf{\text{diag}}}

\global\long\def\mtsign{\mathbf{\text{sign}}}

\global\long\def\mtloss{\mathbf{\text{loss}}}

\global\long\def\mtwhen{\text{when}}

\global\long\def\mtwhere{\text{where}}

\global\long\def\mtif{\text{if}}

\title{Robust Actor-Critic Contextual Bandit for Mobile Health (mHealth)
Interventions}

\author{Feiyun Zhu$^{\dagger}$, Jun Guo$^{\ddagger}$, Ruoyu Li$^{\dagger}$,
Junzhou Huang$^{\dagger}$\thanks{$\dagger$ Department of CSE, University of Texas at Arlington, TX,
76013, USA. $\ddagger$ Department of Statistics, Univeristy of Michigan,
Ann Arbor, MI 48109, USA}}
\maketitle
\begin{abstract}
We consider the actor-critic contextual bandit for the mobile health
(mHealth) intervention. State-of-the-art decision-making algorithms
generally ignore the outliers in the dataset. In this paper, we propose
a novel robust contextual bandit method for the mHealth. It can achieve
the conflicting goal of reducing the influence of outliers, while
seeking for a similar solution compared with the state-of-the-art
contextual bandit methods on the datasets without outliers. Such performance
relies on two technologies: (1) the capped-$\ell_{2}$ norm; (2) a
reliable method to set the thresholding hyper-parameter, which is
inspired by one of the most fundamental techniques in the statistics.
Although the model is non-convex and non-differentiable, we propose
an effective reweighted algorithm and provide solid theoretical analyses.
We prove that the proposed algorithm can find sufficiently decreasing
points after each iteration and finally converges after a finite number
of iterations. Extensive experiment results on two datasets demonstrate
that our method can achieve almost identical results compared with
state-of-the-art contextual bandit methods on the dataset without
outliers, and significantly outperform those state-of-the-art methods
on the badly noised dataset with outliers in a variety of parameter
settings.
\end{abstract}

\section{Introduction}

Due to the explosive growth of smart device (i.e. smartphones and
wearable devices, such as Fitbit etc.) users globally, mobile health
(mHealth) technologies draw increasing interests from the scientist
community\ \cite{PengLiao_2015_Proposal_offPolicyRL,SusanMurphy_2016_CORR_BatchOffPolicyAvgRwd}.
The goal of mHealth is to deliver in-time interventions to device
users, guiding them to lead healthier lives, such as reducing the
alcohol abuse\ \cite{Gustafson_2014_JAMA_drinking,Witkiewitz_2014_JAB_drinkingSmoking}
and increasing physical activities\ \cite{Abby_2013_PlosONE_mobileIntervention}.
With the advanced smart technologies, the mHealth interventions can
be formed according to the users' ongoing statuses and changing needs,
which is more portable and flexible compared with the traditional
treatments. Therefore, mHealth technologies are widely used in lots
of health-related applications, such as eating disorders, alcohol
abuses, mental illness, obesity management and HIV medication adherence\ \cite{SusanMurphy_2016_CORR_BatchOffPolicyAvgRwd,PengLiao_2015_Proposal_offPolicyRL}.

Formally, the tailoring of mHealth intervention is modeled as a sequential
decision-making (SDM) problem. The contextual bandit algorithm provides
a framework for the SDM\ \cite{Ambuj_2017_Springer_FromAds_Interventions}.
In 2014, Lei\ \cite{huitian_2014_NIPS_ActCriticBandit4JITAI} proposed
the first contextual bandit algorithm for the mHealth study. It is
in the actor-critic setting\ \cite{Sutton_2012_MitPress_RLintroduction},
which has two advantages compared with the critic only contextual
bandit methods for the internet advertising\ \cite{YihongLi_2010_WWW_contextualBandit4newsArticleRecommend}:
(a) Lei's method has an explicit parameterized model for the stochastic
policy. By analyzing the estimated parameters in the learned policy,
we could know the key features that contribute most to the policy.
This is important to the behavior scientists for the state (feature)
design. (b) From the perspective of optimization, the actor-critic
algorithm has great properties of quick convergence with low variance\ \cite{Grondman_2012_IEEEts_surveyOfActorCriticRL}.

However, Lei's method assumes that the states at different decision
points are i.i.d. and the current action only influences the immediate
reward\ \cite{huitian_2016_PhdThesis_actCriticAlgorithm}. This assumption
is infeasible in real situations. Taking the delayed effect in the
SDM or mHealth for example, the current action influences not only
the immediate reward but also the next state and through that, all
the subsequent rewards\ \cite{Sutton_2012_MitPress_RLintroduction}.
Accordingly, Lei proposed a new method\ \cite{huitian_2016_PhdThesis_actCriticAlgorithm}
by emphasizing on explorations and reducing exploitations.

Although those two methods serve a good start for the mHealth study,
they assume that the noise in the trajectory follows the Gaussian
distribution. The least square based algorithm is employed to estimate
the expected reward. In reality, however, there are various kinds
of complex noises that can badly degrade the collected data, for example:
(1) the wearable device is unreliable to accurately record the states
and rewards from users under different conditions. (2) The mobile
network is unavailable in some areas. Such case hinders the collecting
of users' states as well as the sending of interventions. (3) The
mHealth relies on the self-reported information (Ecological Momentary
Assessments, i.e. EMAs)\ \cite{Joseph_2016_JPR_EMA} to deliver effective
interventions. However, some users are annoyed at the EMAs. They either
fill out the EMAs via random selections or just leave some or all
the EMAs blank. We consider the various kinds of badly noised observations
in the trajectory as outliers. 

There are several robust methods for the SDM problem\ \cite{lihong_2011_ICML_doublyRobust,zhang_2012_biometircs,huanXu_2009_PhdThesis_robustDecision}.
However, those methods are neither in the actor-critic setting, nor
focusing on the outlier problem. Thus, they are different from this
paper's focus. In the general machine learning task, there are some
robust learning methods to deal with the outlier problem\ \cite{QianSun_2013_SIGKDD_CappedNorm4RPCA,WenhaoJiang_2015_IJCAI_cappedNorm4DictonaryLearning}.
However, none of them are contextual bandit algorithms\textemdash it
may cost of lots of work to transfer their methods to the (actor-critic)
contextual bandit algorithms. Besides, those methods seldom pay attention
to the dataset without outliers. In practice, however, we don't know
whether a given dataset consists of outliers or not. It is necessary
to propose a robust learning method that works well on the dataset
both with and without outliers. 

To alleviate the above problems, we propose a robust contextual bandit
method for the mHealth.  The capped-$\ell_{2}$ norm is used to measure
the learning error for the expected reward estimation (i.e. the critic
updating). It prevents outlier observations from dominating our objective.
Besides, the learned weights in the critic updating are considered
in the actor updating. As a result, the robustness against outliers
are greatly boosted in both actor-critic updatings. There is an important
thresholding parameter $\epsilon$ in our method. We propose a solid
method to set its value according to the distribution of samples,
which is based on one of the most fundamental ideas in the statistics.
It has two benefits: (1) the setting of $\epsilon$ becomes very easy
and reliable; (2) we may achieve the conflicting goal of reducing
the influence of outliers when the dataset indeed contains outliers,
while achieving almost identical results compared with the state-of-the-art
contextual bandit method on the datasets without outliers. Although
the objective is non-convex and non-differentiable, we derive an effective
algorithm. As a theoretical contribution, we prove that our algorithm
could find sufficiently decreasing point after each iteration and
finally converges after a finite number of iterations. Extensive experiment
results on two datasets verify that our methods could achieve clear
gains over the state-of-the-art contextual bandit methods for the
mHealth.

\section{\label{subsec:actorCritic_contextualBandit} Preliminaries}

Multi-armed bandit (MAB) is the simplest algorithm for the sequential
decision making problem (SDM). The contextual bandit is a more practical
extension of the MAB by considering some extra information that is
helpful for the SDM problem\ \cite{Ambuj_2017_Springer_FromAds_Interventions}.
The use of context information allows for many interesting applications,
such as internet advertising and health-care tasks\ \cite{lihong_2011_ICML_doublyRobust,Ambuj_2017_Springer_FromAds_Interventions}. 

In contextual bandit, the expected reward is an core concept that
measures how many rewards we may averagely get when it is in state
$s$ and choosing action $a$, i.e., $\mtexpect\left(r\mid s,a\right)$.
Since the state space is usually very large or even infinite in the
mHealth tasks, the parameterized model is employed to approximate
the expected reward: $\mtexpect\left(r\mid s,a\right)=\mtbfx\left(s,a\right)^{\intercal}\mtbfw$,
where $\mtbfx\left(s,a\right)\in\mtbbR^{u}$ is a feature processing
step that combines information in the state $s$ and the action $a$,
$\mtbfw\in\mtbbR^{u}$ is the unknown coefficient vector. 

In 2014, Lei\ \cite{huitian_2014_NIPS_ActCriticBandit4JITAI} proposed
the first contextual bandit method for the mHealth study. It is in
the actor-critic learning setting. The actor updating is the overall
optimization goal. It aims to learn an optimal policy $\pi^{*}$ that
maximizes the average rewards over all the states and actions\ \cite{Grondman_2012_IEEEts_surveyOfActorCriticRL}.
The objective function is $\pi_{\theta_{n}^{*}}=\arg\max_{\theta_{n}}\widehat{J}\left(\theta_{n}\right)$
for the $n^{\mtth}$ user, where 
\begin{equation}
\widehat{J}\left(\theta_{n}\right)=\sum_{s\in\mtcalS}d_{\text{ref}}^{\left(n\right)}\left(s\right)\sum_{a\in\mtcalA}\pi_{\theta_{n}}\left(a\mid s\right)\mtexpect\left(r\mid s,a\right)\label{eq:obj_actorUpdate_Lei}
\end{equation}
and $d_{\text{ref}}^{\left(n\right)}\left(s\right)$ is a reference
distribution over states for user $n$.

Obviously, we need the estimation of expected rewards $\mtexpect\left(r\mid s,a\right)$
to define the objective\ \eqref{eq:obj_actorUpdate_Lei} for the
actor updating. Such procedure is called the critic updating. State-of-the-art
method generally employs the ridge regression to learn the expected
reward from the observations. The objective is 
\begin{equation}
\min_{\mtbfw_{n}}\sum_{\mtcalU_{i}\in\mtcalD_{n}}\left\Vert \mtbfx\left(s_{i},a_{i}\right)^{\intercal}\mtbfw_{n}-r_{i}\right\Vert _{2}^{2}+\zeta_{c}\left\Vert \mtbfw_{n}\right\Vert _{2}^{2},\label{eq:obj_criticUpdating_current}
\end{equation}
where $\mtcalD_{n}=\left\{ \mtcalU_{i}=\left(s_{i},a_{i,}r_{i}\right)\mid i=0,\cdots,M\right\} $
is the trajectory of observed tuples from the $n^{\mtth}$ user and
$\mtcalU_{i}$ is the $i^{\mtth}$ tuple in $\mtcalD_{n}$; $\zeta_{c}$
is a tuning parameter to control the strength of constraints. It has
a closed-form solution for\ \eqref{eq:obj_criticUpdating_current}
as:
\begin{equation}
\mhbfw_{n}=\left(\mtbfX_{n}\mtbfX_{n}^{\intercal}+\zeta_{c}\mtbfI_{u}\right)^{-1}\mtbfX_{n}\mtbfr_{n}.\label{eq:closedFormSolution_criticUpdating}
\end{equation}
where $\mtbfI_{u}$ is a $u\times u$ identity matrix. Unfortunately,
similar to the existing least square based models in machine learning
and statistics, the objective function in\ \eqref{eq:obj_criticUpdating_current}
is prone to the presence of outliers\ \cite{fpNie_2010_NIPS_JointL21_featureSelection,fyzhu_2014_AAAI_ARSS}. 

\section{Robust Actor-critic Contextual Bandit via the Capped-$\ell_{2}$
norm}

To enhance the robustness in the critic updating, the capped-$\ell_{2}$
norm based measure is used for the estimation of expected rewards.
By imposing the learned weights for the actor updating, we propose
a robust objective for the actor updating.

\subsection{Robust Critic Updating via the Capped-$\ell_{2}$ Norm}

To simplify the notation, we get rid of the subscript index $n$,
which is used to indicate the model for the $n^{\mtth}$ user. The
new objective for the critic updating (i.e. policy evaluation) is
\begin{equation}
\min_{\mtbfw}\sum_{i=1}^{M}\min\left\{ \left\Vert r_{i}-\mtbfx_{i}^{T}\mtbfw\right\Vert _{2}^{2},\epsilon\right\} +\zeta_{c}\left\Vert \mtbfw\right\Vert _{2}^{2},\label{eq:obj_expectedRwd_cappedL2}
\end{equation}
where $\min\left\{ \left\Vert \mtbfy\right\Vert _{2}^{2},\epsilon\right\} $
is the capped-$\ell_{2}$ norm for a vector $\mtbfy$; $\epsilon>0$
is the thresholding hyper-parameter to choose the effective observations
for the critic updating; $\mtbfx_{i}=\mtbfx\left(s_{i},a_{i}\right)$
is the feature for the estimation of expected rewards.

If the residual of the $i^{\mtth}$ tuple is $\left\Vert r_{i}-\mtbfx_{i}^{\intercal}\mtbfw\right\Vert _{2}^{2}>\epsilon$,
 we treat it as an outlier. Its residual is capped to a fixed value
$\epsilon$. That is, the influence of the $i^{\mtth}$ tuple is fixed\ \cite{hongchangGao_2015_CIKM_cappedNorm4NMF},
which can't cause bad influences on the learning procedure. For the
tuples whose residuals satisfy $\left\Vert r_{i}-\mtbfx_{i}^{\intercal}\mtbfw\right\Vert _{2}^{2}\leq\epsilon$,
we consider them as effective observations and keep them as they are
in the optimization process. 

Therefore, it is extremely important to properly set the value of
$\epsilon$. When $\epsilon$ is too large, the outliers that distribute
far away from the majority of tuples will be treated as effective
samples, causing bad influences to the learning procedure. When $\epsilon$
is too small, most tuples are treated as outliers\textemdash there
would be very few of effective samples for the cirtic learning. Such
case easily leads to some unstable policies that contain lots of variances.
Specially if $\epsilon\rightarrow+\infty$, our objective is equivalent
to the least square objective in\ \eqref{eq:obj_criticUpdating_current}. 

As a profound contribution, we propose a reliable method to properly
set the value of $\epsilon$. Our method doesn't need any specific
assumption on the data distribution. It is derivated from the boxplot\textemdash one
of the most fundamental ideas in the statistics\ \cite{dawson_2011_significant_boxplot,williamson_1989_boxplot}.
To give a descriptive illustration of the data distribution, the boxplot
is widely used by specifying 5 points, including the min, lower quartile
$q_{1}$, median, upper quartile $q_{3}$ and max. Based on the 5
points, the boxplot provides a method to detect outliers. Following
this idea, we set the value of $\epsilon$ as 
\begin{equation}
\epsilon=\tau\left(q_{3}+1.5\times IQR\right)\label{eq:outlierDetection_boxplot}
\end{equation}
where $IQR=q_{3}-q_{1}$ is the interquartile range; $\tau$ is introduced
only for the experiment setting \textbf{S2}, otherwise we may ignore
the parameter $\tau$ by keeping it fixed at $1$. Intuitively, the
data points that are $1.5\times IQR$ more above the third quartile
are detected as the outliers. Compared with the state-of-the-art robust
learning methods\ \cite{hongchangGao_2015_CIKM_cappedNorm4NMF,QianSun_2013_SIGKDD_CappedNorm4RPCA,WenhaoJiang_2015_IJCAI_cappedNorm4DictonaryLearning}
in the other fields that have to manually set the thresholding hyper-parameter,
we provide an adaptive method to set $\epsilon$, which is well adapted
to the data distribution.

With the capped-$\ell_{2}$ norm and the method to set $\epsilon$
in\ \eqref{eq:outlierDetection_boxplot}, we can  achieve the conflicting
goals of (a) reducing the influence of outliers when the dataset has
outliers, while (b) seeking for a similar solution compared with that
of the state-of-the-art method if there is no outlier in the dataset.
As a result, our method can deal with various datasets, regardless
of whether they consist of outliers or not. 

\subsection{\label{subsec:problem_transform}Derivation of a General Objective
Function for\ \eqref{eq:obj_expectedRwd_cappedL2}}

To give an efficient algorithm, we consider a more general capped-$\ell_{2}$
norm based objective for\ \eqref{eq:obj_expectedRwd_cappedL2} as
follows
\begin{equation}
\min_{\mtbfx}\sum_{i=1}^{M}\min\left\{ \left\Vert h_{i}\left(\mtbfx\right)\right\Vert _{2}^{2},\epsilon\right\} +\zeta g\left(\mtbfx\right),\label{eq:objGeneral_capL2}
\end{equation}
where $\left\Vert \cdot\right\Vert _{2}^{2}$ is the $\ell_{2}$ norm
for a vector; $h_{i}\left(\mtbfx\right)$ and $g\left(\mtbfx\right)$
are both scalar functions of $\mtbfx$. In this section, we propose
an iteratively re-weighted method to simplify the objective\ \eqref{eq:objGeneral_capL2}.

Due to the non-smooth and non-differentiable property of\ \eqref{eq:objGeneral_capL2},
we could only obtain the sub-gradient of\ \eqref{eq:objGeneral_capL2}
as:
\begin{equation}
\partial\mtcalO\left(\mtbfx\right)=\sum_{i=1}^{M}\partial\min\left(\left\Vert h_{i}\left(\mtbfx\right)\right\Vert _{2}^{2},\epsilon\right)+\zeta\partial g\left(\mtbfx\right).\label{eq:objGeneral_capL2_subGrad}
\end{equation}
 Letting $f=\partial\min\left(\left\Vert h_{i}\left(\mtbfx\right)\right\Vert _{2}^{2},\epsilon\right)$
gives 
\begin{equation}
f=\begin{cases}
0 & \mtif\ \left\Vert h_{i}\left(\mtbfx\right)\right\Vert _{2}^{2}>\epsilon\\
2\left[d_{\text{low}},d_{\text{high}}\right]\partial h_{i}\left(\mtbfx\right) & \mtif\ \left\Vert h_{i}\left(\mtbfx\right)\right\Vert _{2}^{2}=\epsilon\\
2h_{i}\left(\mtbfx\right)\partial h_{i}\left(\mtbfx\right) & \mtif\ \left\Vert h_{i}\left(\mtbfx\right)\right\Vert _{2}^{2}<\epsilon
\end{cases},\label{eq:objGeneral_capL2-term1_subGrad}
\end{equation}
where $d_{\text{low}}=\min\left\{ 0,h_{i}\left(\mtbfx\right)\right\} $
and $d_{\text{high}}=\max\left\{ 0,h_{i}\left(\mtbfx\right)\right\} $.
For the sake of easy optimization, we provide a compact expression
that satisfies the sub-gradient in\ \eqref{eq:objGeneral_capL2-term1_subGrad}
by introducing a variable $u_{i}=1_{\left\{ \left\Vert h_{i}\left(\mtbfx\right)\right\Vert _{2}^{2}<\epsilon\right\} }$.
Then Eq.\ \eqref{eq:objGeneral_capL2_subGrad} is rewritten as 
\begin{equation}
\partial\mtcalO\left(\mtbfx\right)=2\sum_{i}u_{i}h_{i}\left(\mtbfx\right)\partial h_{i}\left(\mtbfx\right)+\zeta\partial g\left(\mtbfx\right).\label{eq:objGeneral_capL2_subGrad-new}
\end{equation}
Since $u_{i}$ depends on $\mtbfx$, it is very challenging to directly
solve the objective\ \eqref{eq:objGeneral_capL2_subGrad-new}. Once
$u_{i}$ is given for every $i,$ the objective\ \eqref{eq:objGeneral_capL2_subGrad}
is equivalent to the following problem 
\begin{equation}
\min_{\mtbfx}\sum_{i=1}^{M}u_{i}\left\Vert h_{i}\left(\mtbfx\right)\right\Vert _{2}^{2}+\zeta g\left(\mtbfx\right)\label{eq:obj_cappedL2_general_simplified}
\end{equation}
in the sense that they have the same partial derivative.

\subsection{Robust Algorithm for the Critic Updating}

In this section, we provide an effective updating rule for the objective
function\ \eqref{eq:obj_expectedRwd_cappedL2} (cf. Proposition\ \ref{prop:updatingRule_for_robustExpectedRwd}
and Algorithm\ \ref{alg:2_actCritic_contextualBandit_4_nUser}).
We prove that our algorithm can find sufficiently decreasing point
after each iteration (cf. Lemma\ \ref{lem:sufficientlyDecreasing})
and finally converge after a finite number of iterations (cf. Theorem\ \ref{thm:stationaryPoints}).
\begin{prop}
\label{prop:updatingRule_for_robustExpectedRwd}The iterative updating
rule\,\eqref{eq:obj_expectedRwd_cappedL2} ($\forall t>1$) is 
\begin{equation}
{\displaystyle \mtbfw^{\left(t\right)}=\left(\mtbfX\mtbfU^{\left(t-1\right)}\mtbfX^{\intercal}+\zeta_{c}\mtbfI\right)^{-1}\mtbfX\mtbfU^{\left(t-1\right)}\mtbfr},\label{eq:updatingRule_4_expectedRwd}
\end{equation}
where $\mtbfU^{\left(t\right)}=\mtdiag\left(\mtbfu^{\left(t\right)}\right)\in\mtbbR^{M\times M}$
is a nonnegative diagonal matrix. The $i^{\mtth}$ element is $u_{i}^{\left(t\right)}=1_{\left\{ \left\Vert r_{i}-\mtbfx_{i}^{\intercal}\mhbfw^{\left(t\right)}\right\Vert _{2}^{2}<\epsilon\right\} }$.
\end{prop}
\begin{lem}
\label{lem:sufficientlyDecreasing}The updating rule in Proposition\ \ref{prop:updatingRule_for_robustExpectedRwd}
leads to sufficient decrease of the objective function $\mtcalO\left(\mtbfw\right)$
in\ \eqref{eq:obj_expectedRwd_cappedL2}\emph{:}
\[
f\left(\mtbfw^{\left(t-1\right)},\mtbfu^{\left(t-1\right)}\right)\geq f\left(\mtbfw^{\left(t\right)},\mtbfu^{\left(t\right)}\right)+\zeta\left\Vert \mtbfw^{\left(t\right)}-\mtbfw^{\left(t-1\right)}\right\Vert _{2}^{2},
\]
where the bivariate function $f\left(\mtbfw,\mtbfu\right)$ is defined
as 
\[
f\left(\mtbfw,\mtbfu\right)=\sum_{i}^{M}u_{i}\left\Vert r_{i}-\mtbfx_{i}^{T}\mtbfw\right\Vert _{2}^{2}+\sum_{i}^{M}\left(1-u_{i}\right)\epsilon+\zeta_{c}\left\Vert \mtbfw\right\Vert _{2}^{2},
\]
which is the same as $\mtcalO\left(\mtbfw\right)$ in\ \eqref{eq:obj_expectedRwd_cappedL2}.

\begin{lem} \label{lem:stickTogether} For $\left\{ \mtbfw^{\left(t\right)},t\geq0\right\} $
in Lemma\ \ref{lem:sufficientlyDecreasing}, we show that $\sum_{t=1}^{\infty}\left\Vert \mtbfw^{\left(t\right)}-\mtbfw^{\left(t-1\right)}\right\Vert _{2}^{2}<\infty$
and consequently 
\[
\lim_{t\rightarrow\infty}\left\Vert \mtbfw^{\left(t\right)}-\mtbfw^{\left(t-1\right)}\right\Vert _{2}^{2}=0.
\]

\end{lem}
\end{lem}
\begin{thm}
\label{thm:stationaryPoints}The updating rule in Proposition\ \ref{prop:updatingRule_for_robustExpectedRwd}
converges after a finite number of iterations.
\end{thm}

\subsection{Robust Actor Updating for the Stochastic Policy}

Besides the critic updating, outliers can also badly influence the
actor updating in\ \eqref{eq:obj_actorUpdate_Lei}, which is our
ultimate objective. To boost its robustness, the estimated weights
learned in the critic updating are considered. Since $d_{\text{ref}}\left(s\right)$
is usually unavailable in reality, the $M$-trial based objective\ \cite{Chou_2014_ACML_PseudoAlg_4_ContextualBandits}
is widely used. Thus, the objective\ \eqref{eq:obj_actorUpdate_Lei}
is rewritten as 
\begin{equation}
\widehat{J}\left(\theta\right)=\frac{u_{i}}{M}\!\sum_{i=1}^{M}\sum_{a\in\mtcalA}\pi_{\theta}\!\left(a\mid s_{i}\right)\mtexpect\!\left(r\mid s_{i},a\right)-\frac{\zeta_{a}}{2}\!\left\Vert \theta\right\Vert _{2}^{2},\label{eq:obj_actorUpdate_our}
\end{equation}
where $\mtexpect\left(r\mid s_{i},a\right)=\mtbfx\left(s_{i},a\right)^{\intercal}\mhbfw$
is the estimated expected reward; $\left\Vert \theta\right\Vert _{2}^{2}$
is the least square constraint to make the objective\ \eqref{eq:obj_actorUpdate_our}
a well-posed problem and $\zeta_{a}$ is a balancing parameter that
controls the penalization strength\ \cite{huitian_2014_NIPS_ActCriticBandit4JITAI}.

Compared with the current objective for the actor updating in\ \eqref{eq:obj_actorUpdate_Lei},
our objective has an extra weight term $\left\{ u_{i}\right\} _{i=1}^{M}$,
which gives those tuples, whose residuals are very large in the critic
updating, zero weights. As a result, the outlier tuples that are far
away from the majority of tuples are removed from the actor updating,
enhancing the robustness. The algorithm of the actor updating performs
the maximization of\ \eqref{eq:obj_actorUpdate_our} over $\theta$.
This is learned via the Sequential Quadratic Programming (SQP) algorithm.
We utilize the implementation of SQP with finite-difference approximation
to the gradient in the \emph{fmincon} function of Matlab. 
\begin{algorithm}[t]
\caption{robust actor-critic contextual bandit for user $n$.\label{alg:2_actCritic_contextualBandit_4_nUser}}
 \textbf{Input}: $\zeta_{a},\zeta_{c},\epsilon,\mtbfu=\left[1,\cdots,1\right]\in\mtbbR^{T}$

\begin{algorithmic}[1] 

\STATE Initialize states $s_{0}\!\in\!\mtbbR^{p}$ and policy parameters
$\theta_{0}\!\in\!\mtbbR^{m}$. 

\REPEAT 

\STATE \emph{/{*}}\textbf{\emph{Critic updating}}\emph{ for the expected
reward{*}/}

\REPEAT

\STATE Update $\mbox{\ensuremath{\mhbfw}\ for the expected reward via\ }$\eqref{eq:updatingRule_4_expectedRwd}.

\STATE Update the weights $\mtbfu$ via the Proposition\ \ref{prop:updatingRule_for_robustExpectedRwd}.

\UNTIL{ convergence }

\STATE \emph{/{*}}\textbf{\emph{Actor updating}}{*}/ via $\widehat{\theta}=\arg\max_{\theta}\widehat{J}\left(\theta\right)$,
where
\[
\widehat{J}\left(\theta\right)=\frac{u_{i}}{M}\sum_{i=1}^{M}\sum_{a\in\mtcalA}\pi_{\theta}\left(a\mid s_{i}\right)\mtexpect\left(r\mid s_{i},a\right)-\frac{\zeta_{a}}{2}\left\Vert \theta\right\Vert _{2}^{2}.
\]
\UNTIL{ convergence }

\end{algorithmic} 

\textbf{Output}: the stochastic policy for $n^{\mtth}$ user, i.e.
$\pi_{\widehat{\theta}_{n}}\left(a\mid s\right)$. 
\end{algorithm}

\section{Experiment}

\subsection{\label{subsec:Datasets}Two Datasets}

Two datasets are used to verify the performance of our method. The
first dataset is on the personalizing treatment delivery in mobile
health, which is a common application of the sequential decision making
algorithm\ \cite{SusanMurphy_2016_CORR_BatchOffPolicyAvgRwd}. In
this paper, we focus on the Heartsteps, where the participants are
periodically sent activity suggestions aimed at decreasing sedentary
behavior\ \cite{Klasnja_2015_APA_HeartSteps}. Specifically, Heartsteps
is a 42-days trial study where 50 participants are involved. For each
participant, there are 210 decision points\textemdash five decisions
per participant per day. At each time point, the set of intervention
actions can be the intervention type, as well as whether or not to
send interventions. The intervention actions generally depend on the
state of the participant as well as the formerly sent interventions.
Interventions can be sent via smartphones, or via other wearable devices
like a wristband\ \cite{Walter_2015_Significance_RandomTrialForFitbitGeneration}.
\begin{table}[t]
\begin{centering}
\caption{The ElrAR of nine contextual methods on the two datasets: Heartsteps
and chain walk. (experiment setting\textbf{ S1})\label{tab:_ExpSetting_S1}}
\includegraphics[width=0.99\columnwidth]{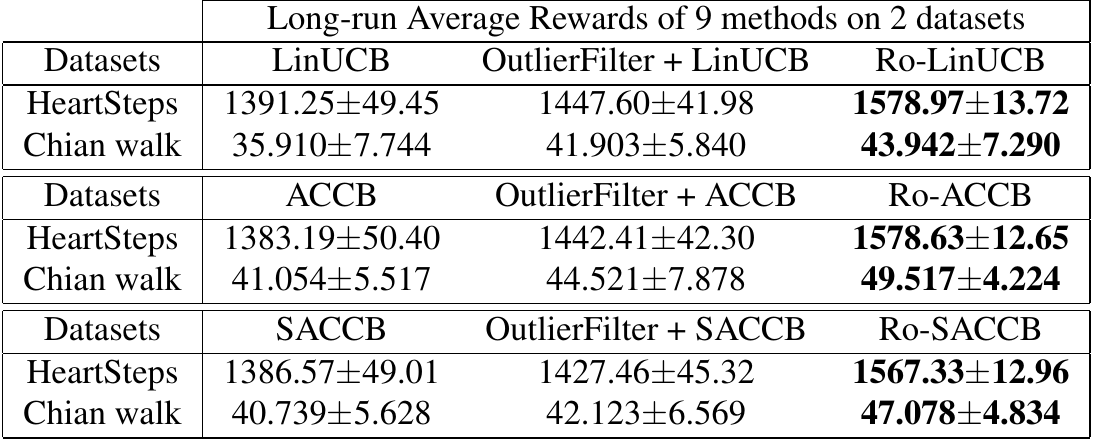}
\par\end{centering}
\centering{}\vspace{-0.25cm}
\end{table}

The Heartsteps is a common application for the contextual bandit algorithm\ \cite{huitian_2016_PhdThesis_actCriticAlgorithm,SusanMurphy_2016_CORR_BatchOffPolicyAvgRwd}.
The goal is to learn an optimal SDM algorithms to decide what type
of intervention actions to send to each user to maximize the cumulative
steps each user takes. The resulting data for each participant is
a sequence $\mtcalD=\left\{ s_{0},a_{0,}r_{0},\cdots,s_{209},a_{209},r_{209},r_{210},s_{210}\right\} $,
where $s_{t}$ is the participant's state at time $t$. It is a three
dimensional vector that consists of (1) the weather condition, (2)
the engagement of participants, (3) the treatment fatigue of participants.
$a_{t}$ indicates whether or not to send the intervention to users;
since the goal of Heartsteps is to increase the participant\textquoteright s
activities, we define the reward, $r_{t}$, as the step count for
the 3 hours following a decision point.

The second dataset is the 4-state chain walk, which is a benchmark
dataset for the (contextual) bandit and reinforcement learning study.
Please refer to\ \cite{Michail_2003_JMLR_LSPI_LSTDQ} for the details
of the chain walk dataset. The reward vector over the four states
is $\left(0,100,100,0\right)$ in this paper. 
\begin{table*}[t]
\begin{centering}
\caption{The ElrAR of six methods \textbf{\emph{vs.}} outlier ratio $\psi$
on two datasets: (1) Heartsteps and (2) chain walk. (experiment\textbf{
S2}). \label{tab:_ExpSetting_S3}}
\includegraphics[width=1.75\columnwidth]{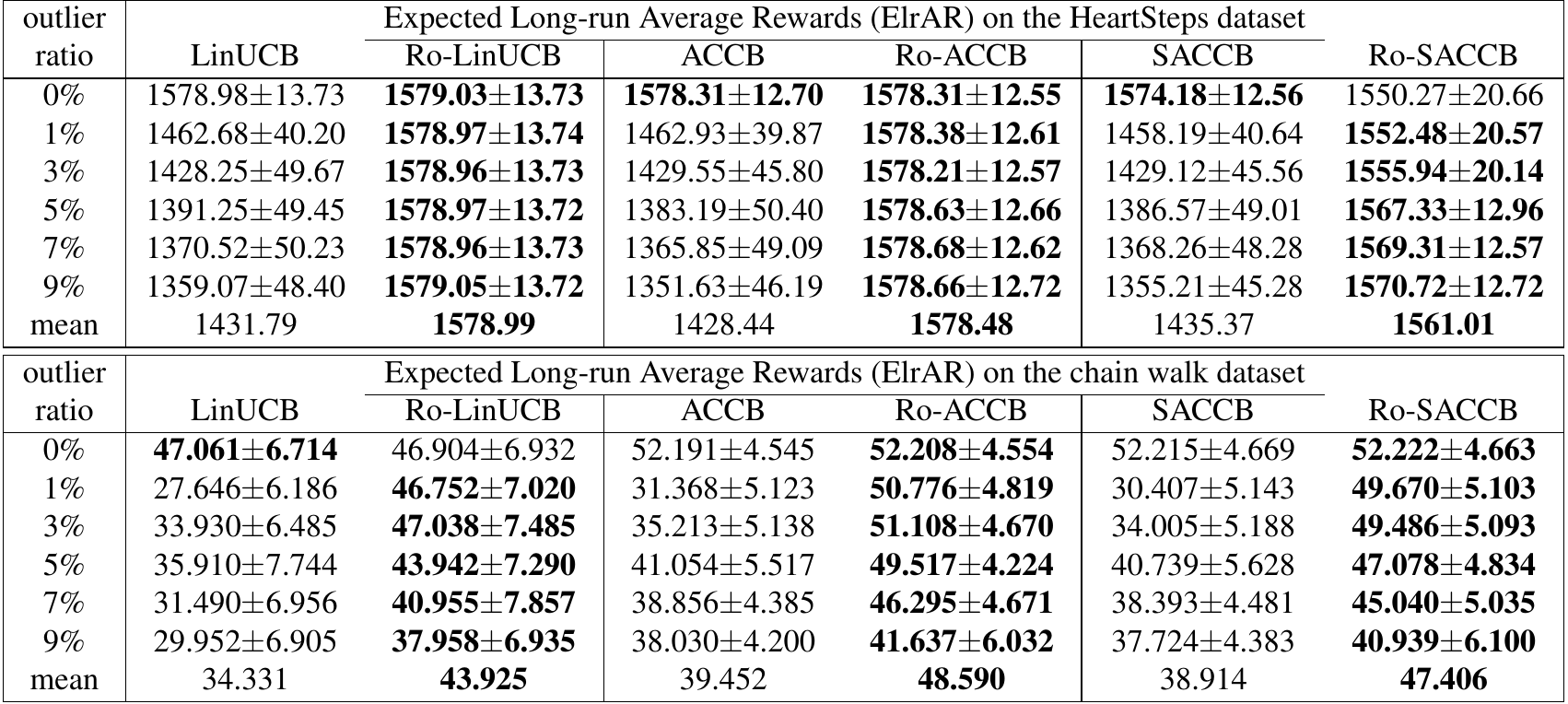}
\par\end{centering}
\begin{centering}
\vspace{-0.25cm} \caption{The ElrAR of six methods \textbf{\emph{vs.}} outlier strength $\nu$
on two datasets: (1) Heartsteps and (2) chain walk. (experiment\textbf{
S3}). \label{tab:_ExpSetting_S4}}
\par\end{centering}
\begin{centering}
\includegraphics[width=1.75\columnwidth]{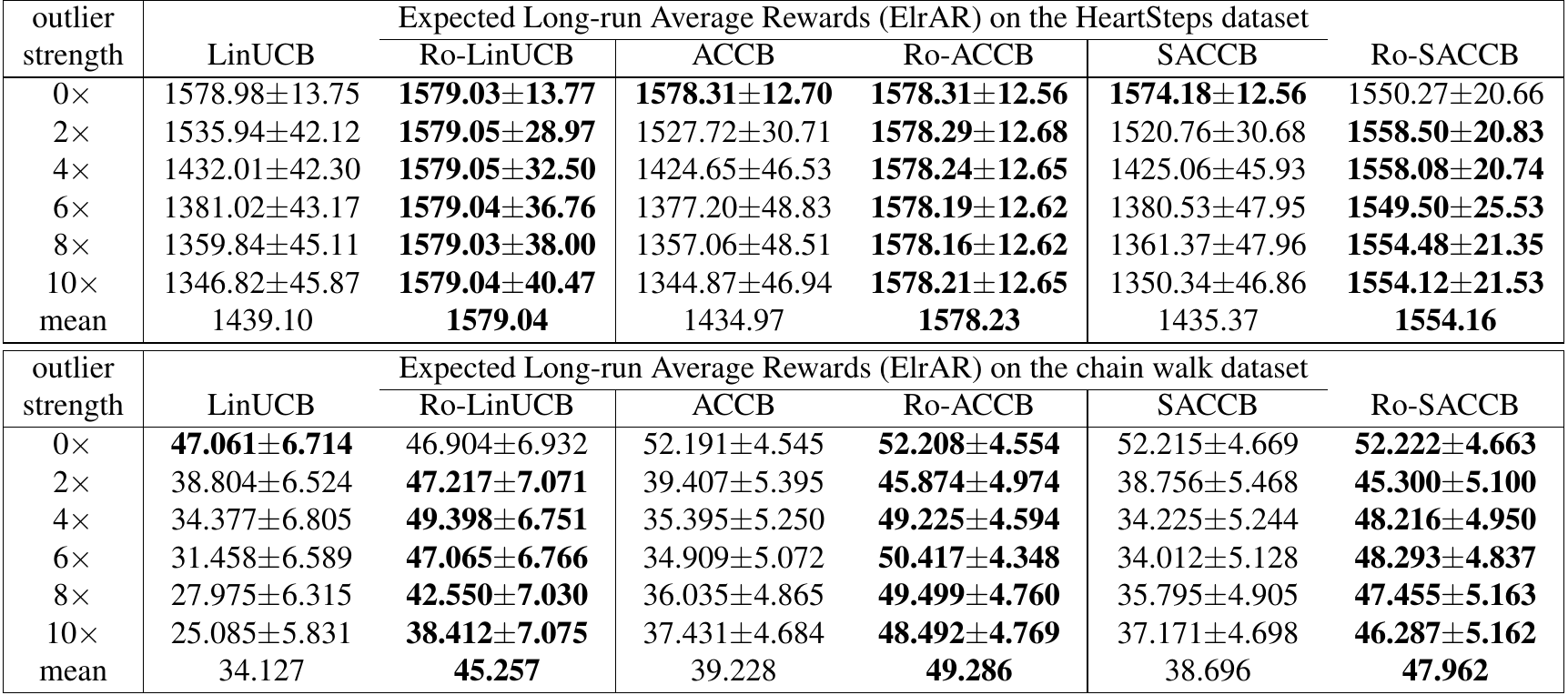} 
\par\end{centering}
\centering{}\vspace{-0.35cm}
\end{table*}

\subsection{\label{subsec:ComparedMethods_ParameterSetting}Nine Compared Methods}

There are nine contextual bandit methods compared in the experiment,
including three state-of-the-art methods: (1) Linear Upper Confidence
Bound Algorithm (LinUCB) is one of the most famous contextual bandit
methods used in the personalized news recommendation\ \cite{YihongLi_2010_WWW_contextualBandit4newsArticleRecommend,Chunyen_2015_AAAI_CBBL};
(2) the actor-critic contextual bandit (ACCB) is the first SDM method
for the mHealth intervention\ \cite{huitian_2014_NIPS_ActCriticBandit4JITAI};
(3) the stochasticity constrained actor-critic contextual bandit (SACCB)
for the mHealth\ \cite{huitian_2016_PhdThesis_actCriticAlgorithm}.
We improve the above three methods by first using the state-of-the-art
outlier filter\ \cite{Hancong_2004_CCE,Gustafson_2014_JAMA_drinking}
to remove outliers, then employing the above three contextual bandit
algorithms for the SDM task. In the $7^{\mtth}$ to $9^{\mtth}$ compared
methods, we apply the proposed robust model and optimization algorithms
in the three state-of-the-art contextual bandit methods, leading to
(7) Robust LinUCB (Ro-LinUCB for short), (8) Robust ACCB (Ro-ACCB),
(9) Robust SACCB (Ro-SACCB). 

\subsection{Evaluation Methodology and Parameter Setting}

It has been a challenging problem to reliably evaluate the sequential
decision making (e.g., bandit and reinforcement learning) algorithms
when the simulator (like the Atari games) is unavailable\ \cite{YihongLi_2010_WWW_contextualBandit4newsArticleRecommend,lihongli_2015_ArXiv_RRL}.
After the model is trained, we use it to interact with the enviroment
(or simulator) to collect thousands of immediate rewards for the calculation
of the long term rewards as the evaluation metric. However for a wide
variety of applications including our HeartSteps, the simulator is
unavailable. In this paper, we use a benchmark evalutaion methods\ \cite{lihongli_2015_ArXiv_RRL}.
The main idea is to make use of the collected dataset to build a simulator,
based on which we train and evaluate the contextual bandit methods.
Such case makes it impossible to train and evaluate the contextual
bandit algorithms on up to ten datasets like the general supervised
learning tasks. 
\begin{figure*}
\centering{}\includegraphics[width=0.96\linewidth]{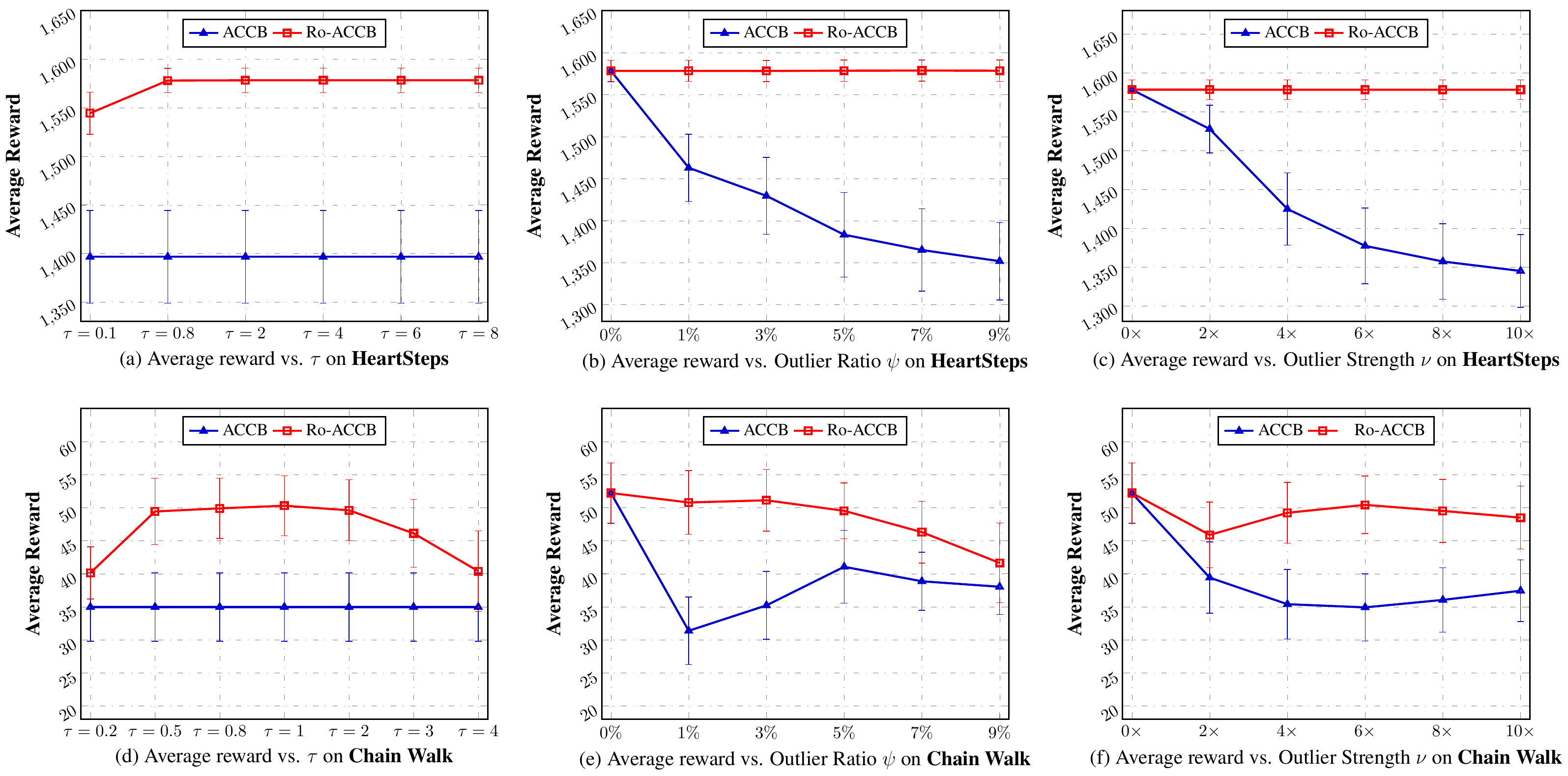}\vspace{-0.35cm}\caption{The ElrAR of two contextual bandit methods \textbf{\emph{vs.}} $\epsilon$
(or $\tau$ in Eq.\ \eqref{eq:outlierDetection_boxplot}), outlier
ratio $\psi$ and outlier strength $\nu$ respectively on the two
datasets. The top row of sub-figures illustrate the results on the
HeartSteps, and the bottom row shows the results on the chain walk.
The three columns of sub-figures show the result in the experiment
setting (\textbf{S4}), (\textbf{S2}) and (\textbf{S3}) respectively.
\label{fig:P_vs_3Settings}}
\vspace{-0.25cm}
\end{figure*}

In the Heartsteps study, there are 50 users; the simulator for each
user is as follows: the initial state is drawn from the Gaussian distribution
$S_{0}\sim\mtcalN_{p}\left\lbrace 0,\Sigma\right\rbrace $, where
$\Sigma$ is a $p\times p$ covariance matrix with pre-defined elements.
For $t\geq0,$ $a_{t}$ is drawn from the learned policy $\pi_{\widehat{\theta}_{n}}\left(a_{t}\mid s_{t}\right)$
during the evaluation procedure. When $t\geq1$, the state and immediate
reward are generated as 
\begin{align}
S_{t,1}=\  & \beta_{1}S_{t-1,1}+\xi_{t,1},\nonumber \\
S_{t,2}=\  & \beta_{2}S_{t-1,2}+\beta_{3}A_{t-1}+\xi_{t,2},\label{eq:Dat=0000231_stateTrans_cmp3}\\
S_{t,3}=\  & \beta_{4}S_{t-1,3}+\beta_{5}S_{t-1,3}A_{t-1}+\beta_{6}A_{t-1}+\xi_{t,3},\nonumber \\
R_{t}=\  & \beta_{13}\times[\beta_{7}+A_{t}\times\left(\beta_{8}+\beta_{9}S_{t,1}+\beta_{10}S_{t,2}\right)\label{eq:Dat=0000231_ImmediateRwd_cmp3}\\
 & +\beta_{11}S_{t,1}-\beta_{12}S_{t,3}+\varrho_{t}],\nonumber 
\end{align}
where $\bm{\beta}\!=\!\left\{ \beta_{1},\!\cdots\!,\beta_{13}\right\} $={[}0.4,0.3,0.4,0.7,0.05,0.6,3,0.25,0.25
,0.4,0.1,0.5,500{]} is the main parameter for the MDP system. $\left\{ \xi_{t,i}\right\} _{i=1}^{3}\!\sim\!\mtcalN\left(0,1\right)$
and $\varrho_{t}\!\sim\!\mtcalN\left(0,9\right)$ are the gaussian
noises in the state\ \eqref{eq:Dat=0000231_stateTrans_cmp3} and
in the reward\ \eqref{eq:Dat=0000231_ImmediateRwd_cmp3} respectively.

The parameterized policy is assumed to follow the Boltzmann distribution
$\pi_{\theta}\left(a\mid s\right)\!=\!\frac{\exp\left[-\theta^{\intercal}\phi\left(s,a\right)\right]}{\sum_{a'}\exp\left[-\theta^{\intercal}\phi\left(s,a'\right)\right]}$,
where $\phi\left(s,a\right)=\left[as^{\intercal},a\right]^{\intercal}\in\mtbbR^{4}$
is the policy feature, $\theta\in\mtbbR^{4}$ is the unknown coefficient
vector. The feature for the estimation of expected rewards is set
$\mtbfx\left(s,a\right)=\left[1,s^{\intercal},a,s^{\intercal}a\right]^{\intercal}\in\mtbbR^{8}$.
The $\ell_{2}$ constraint for the actor-critic learning is set as
$\zeta_{a}=\zeta_{c}=0.001$. The outlier ratio and strength are set
$\psi\!=\!4\%$ and $\nu\!=\!5$ respectively. In our methods, $\tau$
is set as 1 by default. 

The expected long-run average reward (ElrAR)\ \cite{SusanMurphy_2016_CORR_BatchOffPolicyAvgRwd}
is used to quantify the quality of the estimated policy $\pi_{\widehat{\theta}_{n}}$
for $n\in\left\{ 1,\cdots,50\right\} $. Intuitively, ElrAR measures
the average steps users take per day in the long-run Heartsteps study
when we use the estimated policy $\pi_{\widehat{\theta}_{n}}$ to
send interventions to users. There are two steps to obtain the ElrAR:
(a) get the average reward $\eta^{\pi_{\widehat{\theta}_{n}}}$ for
the $n^{\mtth}$ user by averaging the rewards over the last $4,000$
decision points in a trajectory of $5,000$ tuples under the policy
$\pi_{\widehat{\theta}_{n}}$; (b) the ElrAR $\mathbb{E}\left[\eta^{\pi_{\hat{\theta}}}\right]$
is achieved by averaging over the $50$ $\eta^{\pi_{\widehat{\theta}_{n}}}$s. 

\subsection{Comparisons in the Four Experiment Settings}

We carry out the following experiments to verify four aspects of the
contextual bandit methods: 

(\textbf{S1}) To verify the significance of the proposed method, we
compare nine contextual bandit methods on two datases: (1) HeartSteps
and (2) chain walk. The experiment results are summarized in Table\ \ref{tab:_ExpSetting_S1},
where there are three sub-tables; each sub-table displays three methods
in a type: (a) the state-of-the-art contextual bandit, like ACCB;
(b) ``OutlierFilter + ACCB'' means that we first use the state-of-the-art
outlier filter\ \cite{Hancong_2004_CCE,Suomela_2014_ArXiv_medianFiltering}
to get rid of the outliers, then employ the state-of-the-art contextual
bandit method ACCB for the SDM task; (c) is the proposed robust contextual
bandit method (Ro-ACCB). As we shall see, ``OutlierFilter'' is helpful
to improve the performance of the state-of-the-art contextual bandit
methods. However, our three methods (i.e., Ro-LinUCB, Ro-ACCB and
Ro-SACCB) always obtain the best results compared with all the other
state-of-the-art methods in their type on both datasets. Compared
with the best state-of-the-art method, our three methods improve 131.4,
136.2 and 139.9 steps respectively on the HeartSteps. Although there
are lots of general outlier detection or outlier filter methods that
can be helpful to relieve the bad influence of outliers, it is still
meaningful to specifically propose a robust contextual bandit algorithm. 

(\textbf{S2}) In this part, the ratio of tuples $\psi$ that contains
outliers rises from $0\%$ to $9\%$. The experiment results are summarized
in Table\ \ref{tab:_ExpSetting_S3} and Figs.\ \ref{fig:P_vs_3Settings}b,\ \ref{fig:P_vs_3Settings}e.
In Table\ \ref{tab:_ExpSetting_S3}, there are two sub-tables, displaying
the ElrARs on the HeartSteps in the top, and the ElrARs on the chain
walk in the bottom. As we can see, when $\mu=0\%$, there is no outlier
in the trajectory. In such case, our results are (almost) identical
to that of LinUCB, ACCB and SACCB on both datasets. When $\psi$ rises,
our results keep stable, while the ElrARs, of LinUCB, ACCB and SACCB,
decrease dramatically. Compared with ACCB, Ro-ACCB averagely achieves
an improvement of $10.5\%$ on the HeartSteps dataset and $23.2\%$
on the chain walk dataset. Such results demonstrate that our method
is able to deal with the badly noised dataset that consists of a large
percentage (up to $9\%$) of outliers.

(\textbf{S3}) In this part, the strength of outliers $\nu$ ranges
from $0$ to $10$ times of the average value in the trajectory. The
experiment results are summarized in Table\ \ref{tab:_ExpSetting_S4},
and Figs.\ \ref{fig:P_vs_3Settings}c,\ \ref{fig:P_vs_3Settings}f.
We have three observations based on the experiment results: (1) when
$\nu=0$, there is no outlier in the trajectory. Out methods achieve
similar results with that of LinUCB, ACCB and SACCB; (2) when $\nu$
rises in the domain, our method keeps stable on the HeartSteps and
decreases slightly on the chain walk. However, the results of LinUCB,
ACCB and SACCB decrease obviously when $\nu$ rises. Such phenomena
verify that our method is able to deal with the dataset both with
or without outliers. Besides, we may use our method on the dataset
with various strengths of outliers. 

(\textbf{S4}) In this part, the value of $\tau$ ranges from $0.1$
to $8$ on the HeartSteps dataset and from $0.2$ to $4$ on the chain
walk dataset. The experiment results are displayed in Figs.\ \ref{fig:P_vs_3Settings}a
and\ \ref{fig:P_vs_3Settings}d. As we shall see, the proposed method
obtains clear advantage over the state-of-the-art method, i.e., ACCB\ \cite{huitian_2014_NIPS_ActCriticBandit4JITAI},
in a wide range of $\tau$ settings. In average, our method improves
the ElrAR by $12.6\%$ on the HeartSteps and $33.1\%$ on the chain
walk, compared with the ACCB. Such results verify that the proposed
method to set $\tau$ is very promising. It is able to adapt to the
data property and select the effective, neither too few nor too many,
tuples in the trajectory for the actor-critic updating. Note that
ACCB does not have the parameter $\tau$. Thus the result of ACCB
remains unchanged as $\tau$ rises.

\section{Conclusion and Discussion}

To deal with the outlier in the trajectory, we propose a robust actor-critic
contextual bandit for the mHealth intervention. The capped-$\ell_{2}$
norm is employed to boost the robustness for the critic updating.
With the learned weights in the critic updating, we propose a new
objective for the actor updating, enhancing its robustness. Besides,
we propose a solid method to set an important thresholding parameter
in the capped-$\ell_{2}$ norm. With it, we can achieve the conflicting
goal of boosting the robustness of our algorithm on the dataset with
outliers, and achieving almost identical results compared with the
state-of-the-art method on the datasets without outliers. Besides,
we provide theoretical guarantee for our algorithm. It shows that
our algorithm could find sufficiently decreasing point after each
iteration and finally converges after a finite number of iterations.
Extensive experiment results show that in a variety of parameter settings
our method achieves significant improvements. \vspace{-0.15cm}

\subsection*{Appendix 1: the proof of Proposition\ \ref{prop:updatingRule_for_robustExpectedRwd} }
\begin{proof}
According to the analyses in Eqs\ \eqref{eq:objGeneral_capL2} and\ \eqref{eq:obj_cappedL2_general_simplified},
we simplify\ \eqref{eq:obj_expectedRwd_cappedL2} into the following
objective \vspace{-0.15cm}
\begin{equation}
\min_{\mtbfw}\ \sum_{i=1}^{M}u_{i}\left\Vert r_{i}-\mtbfx_{i}^{\intercal}\mtbfw\right\Vert _{2}^{2}+\zeta_{c}\left\Vert \mtbfw\right\Vert _{2}^{2}.\label{eq:obj_cappedL2_general_simplified-1}
\end{equation}
Taking the partial derivative and setting it to zero give us the
updating rule \vspace{-0.15cm} 
\[
\mhbfw^{\left(t\right)}=\left(\mtbfX\mtbfU^{\left(t-1\right)}\mtbfX^{\intercal}+\zeta\mtbfI\right)^{-1}\mtbfX\mtbfU^{\left(t-1\right)}\mtbfr,
\]
where $\mtbfU^{\left(t\right)}=\mtdiag\left(\mtbfu^{\left(t\right)}\right)\in\mtbbR^{M\times M}$
is nonnegative diagonal. The $i^{\mtth}$ element is $u_{i}^{\left(t\right)}=1_{\left\{ \left\Vert r_{i}-\mtbfx_{i}^{\intercal}\mhbfw^{\left(t\right)}\right\Vert _{2}^{2}<\epsilon\right\} }.$
\end{proof}

\subsection*{Appendix 2: the proof of Lemma\ \ref{lem:sufficientlyDecreasing}}
\begin{proof}
For $t\geq1,$ when fix $\mtbfu^{\left(t-1\right)}$, we find that
\begin{align*}
f\left(\mtbfw,\mtbfu^{\left(t-1\right)}\right)= & \sum_{i}^{M}u_{i}^{\left(t-1\right)}\left\Vert r_{i}-\mtbfx_{i}^{T}\mtbfw\right\Vert _{2}^{2}+\\
 & \sum_{i}^{M}\left(1-u_{i}^{\left(t-1\right)}\right)\epsilon+\zeta_{c}\left\Vert \mtbfw\right\Vert _{2}^{2}
\end{align*}
is a quadratic function, which is strongly convex. The updating rule
in Proposition\ \ref{prop:updatingRule_for_robustExpectedRwd} minimizes
$f\left(\mtbfw,\mtbfu^{\left(t-1\right)}\right)$ globally over $\mtbfw.$
Via the strong convexity of $f\left(\mtbfw,\mtbfu^{\left(t-1\right)}\right)$,
we have
\begin{align}
 & f\left(\mtbfw^{\left(t-1\right)},\mtbfu^{\left(t-1\right)}\right)\nonumber \\
\geq & f\left(\mtbfw^{\left(t\right)},\mtbfu^{\left(t-1\right)}\right)+\zeta_{c}\left\Vert \mtbfw^{\left(t\right)}-\mtbfw^{\left(t-1\right)}\right\Vert _{2}^{2}.\label{eq:fix_u_strongConvex}
\end{align}
 When fixing $\mtbfw^{\left(t\right)}$, updating $\mtbfu^{\left(t\right)}$
gives 
\begin{align}
 & f\left(\mtbfw^{\left(t\right)},\mtbfu^{\left(t\right)}\right)-f\left(\mtbfw^{\left(t\right)},\mtbfu^{\left(t-1\right)}\right)\nonumber \\
= & -\sum_{i}^{M}\left|\left\Vert r_{i}-\mtbfx_{i}^{\intercal}\mtbfw\right\Vert _{2}^{2}-\epsilon\right|1_{\left\{ u_{i}^{\left(t\right)}\neq u_{i}^{\left(t-1\right)}\right\} }\leq0.
\end{align}
Finally $\forall t\geq1$, we conclude the following inequation 
\begin{align}
 & f\left(\mtbfw^{\left(t-1\right)},\mtbfu^{\left(t-1\right)}\right)\nonumber \\
\geq & f\left(\mtbfw^{\left(t\right)},\mtbfu^{\left(t\right)}\right)+\zeta_{c}\left\Vert \mtbfw^{\left(t\right)}-\mtbfw^{\left(t-1\right)}\right\Vert _{2}^{2}.
\end{align}
\end{proof}

\subsection*{Appendix 3: the proof of Lemma\ \ref{lem:stickTogether}}
\begin{proof}
We sum up the function descent inequality\ \eqref{eq:fix_u_strongConvex}
for $t=1,2,\cdots,T$: 
\begin{align}
 & \sum_{t=1}^{T}\left\Vert \mtbfw^{\left(t\right)}-\mtbfw^{\left(t-1\right)}\right\Vert _{2}^{2}\nonumber \\
\leq & \frac{2}{\zeta_{c}}\sum_{i=1}^{T}\left[O\left(\mtbfw^{\left(t-1\right)}\right)-O\left(\mtbfw^{\left(t\right)}\right)\right]\nonumber \\
= & \frac{2}{\zeta_{c}}\left[O\left(\mtbfw^{\left(0\right)}\right)-O\left(\mtbfw^{\left(T\right)}\right)\right].\label{eq:proof_of_Lemma3}
\end{align}
From\ \eqref{eq:fix_u_strongConvex}, the sequence $\left\{ O\left(\mtbfw^{\left(t\right)}\right),t\geq0\right\} $
is nonincreasing with $O\left(\mtbfw\right)\geq0$, $\forall\mtbfw\in\mtcalW$.
Taking the limit of $T\rightarrow\infty$ on both sides of\ \eqref{eq:proof_of_Lemma3},
we get 
\[
\sum_{t=1}^{T}\left\Vert \mtbfw^{\left(t\right)}-\mtbfw^{\left(t-1\right)}\right\Vert _{2}^{2}\leq\infty
\]
 and thus
\[
\lim_{t\rightarrow\infty}\left\Vert \mtbfw^{\left(t\right)}-\mtbfw^{\left(t-1\right)}\right\Vert _{2}^{2}=0.
\]
\end{proof}
\begin{rem}
With Lemma\ \ref{lem:sufficientlyDecreasing} and Lemma\ \ref{lem:stickTogether},
one can actually show that, given a fixed outlier thresholding $\epsilon>0$,
the algorithm converges after \emph{finite number} of iterations.
\end{rem}

\subsection*{Appendix 4: the proof of Theorem\ \ref{thm:stationaryPoints}}
\begin{proof}
We first show that the sequence $\left\{ O\left(\mtbfw^{\left(t\right)}\right),t\in\mtbbN\right\} $
is bounded. It is easily to see that $h_{i}\left(\mtbfw\right)=\left\Vert r_{i}-\mtbfx_{i}^{\intercal}\mtbfw\right\Vert _{2}^{2}$
maps an unbounded set to an unbounded range. If 
\[
\min_{i=1,\cdots,M}\left\{ \left\Vert r_{i}-\mtbfx_{i}^{\intercal}\mtbfw^{\left(\tau\right)}\right\Vert _{2}^{2}\right\} >\epsilon,
\]
the critic update (the updating rule in Proposition\ \ref{prop:updatingRule_for_robustExpectedRwd})
will stop with $\mtbfw^{\left(t\right)}=\mtbfzero$, $\forall t\geq\tau$.
So the sequence $\left\{ O\left(\mtbfw^{\left(t\right)}\right),t\in\mtbbN\right\} $
must be bounded such that $\left\Vert \mtbfw^{\left(t\right)}\right\Vert _{2}^{2}\leq B$,
for some $B>0$. 

Now $\forall t\ge1$, we have 
\begin{align*}
 & \left|\left\Vert r_{i}-\mtbfx_{i}^{\intercal}\mtbfw^{\left(t\right)}\right\Vert _{2}^{2}-\left\Vert r_{i}-\mtbfx_{i}^{\intercal}\mtbfw^{\left(t-1\right)}\right\Vert _{2}^{2}\right|\\
\leq & B'\left\Vert \mtbfw^{\left(t\right)}-\mtbfw^{\left(t-1\right)}\right\Vert _{2}^{2}
\end{align*}
for some $B'>0$. Then via Lemma\ \ref{lem:stickTogether}, for a
given fixed outlier threshold parameter $\epsilon>0$, we deduce that
there exists $T\in\mtbbN$ when $t\geq T$, we have 
\[
\left(\left\Vert r_{i}-\mtbfx_{i}^{\intercal}\mtbfw^{\left(t\right)}\right\Vert _{2}^{2}-\epsilon\right)\cdot\left(\left\Vert r_{i}-\mtbfx_{i}^{\intercal}\mtbfw^{\left(t-1\right)}\right\Vert _{2}^{2}-\epsilon\right)\geq0,
\]
 $\forall i=1,\cdots,T$. That is $\forall i=1,\cdots,T$, $u_{i}$
remain unchanged for all $t\geq T$ and the problem will become a
least square problem. Thus after $T<\infty$ steps, the updating rule
in Proposition\ \ref{prop:updatingRule_for_robustExpectedRwd} will
converge at a closed form solution
\[
\mhbfw^{\left(t\right)}=\left(\mtbfX\mtbfU^{\left(t-1\right)}\mtbfX^{\intercal}+\zeta\mtbfI\right)^{-1}\mtbfX\mtbfU^{\left(t-1\right)}\mtbfr.
\]
and its corresponding $\mtbfU^{\left(t\right)}$.
\end{proof}
{\small{}\bibliographystyle{20_home_fyzhu_DATA_Dropbox_self_Folder_myWorksO___oxs_201702_IJCAI_RobustContextualBanit_aaai}
\bibliography{19_home_fyzhu_DATA_Dropbox_self_Folder_myWorksO___I_RobustContextualBanit_referenceBib_AAAI18}
}{\small \par}
\end{document}